%% file: iclr2025_conference.tex
\documentclass{article} % For LaTeX2e
\usepackage{iclr2025_conference,times}

\input{math_commands.tex}

\usepackage{hyperref}
\usepackage{url}
\usepackage{graphicx}
\usepackage{multirow}
\usepackage{rotating}
\title{Sequential Compression Layers for Efficient Federated Learning in Foundational Models}

\author{Navyansh Mahla  \\
Indian Institute of Technology Bombay\\
Mumbai, Maharashtra, India \\
\texttt{navyanshmahla17@gmail.com} \\
\And
Sunny Gupta \\
Indian Institute of Technology Bombay \\
Mumbai, Maharashtra, India \\
\texttt{sunnygupta@iitb.ac.in} \\
\AND
Amit Sethi \\
Indian Institute of Technology Bombay \\
Mumbai, Maharashtra, India \\
\texttt{asethi@iitb.ac.in}
}

\iclrfinalcopy % Uncomment for camera-ready version, but NOT for submission.
\begin{document}

\maketitle

\begin{abstract}
Federated Learning (FL) has emerged as a promising paradigm for fine-tuning large language models (LLMs) across distributed nodes, each containing private, decentralized data. While LoRA has become the de facto standard for parameter-efficient federated fine-tuning, recent theoretical and empirical analyses reveal fundamental limitations in its performance within the federated learning context. We propose a simple yet effective parameter-efficient fine-tuning method that surpasses LoRA-based approaches. Our method introduces a lightweight multi-layer perceptron (MLP) between the existing \textit{up\_proj} and \textit{down\_proj} layers within the feed-forward network of transformer blocks. Through extensive experiments, we demonstrate that our approach consistently outperforms state-of-the-art LoRA-based methods across various tasks in both language and vision modalities. While privacy considerations remain central to federated learning research, our work focuses specifically on analyzing the convergence behavior and final model performance across different federation strategies, providing empirical evidence for the better performance of our proposed architecture in distributed fine-tuning scenarios.
\end{abstract}

\section{Introduction}
Large Language Models (LLMs) have revolutionized Natural Language Processing (NLP), demonstrating remarkable capabilities across diverse tasks. Fine-tuning these models on domain-specific datasets has become crucial for optimizing their performance on specialized downstream tasks. However, real-world data is often distributed across institutions and cannot be centralized due to privacy concerns, regulatory requirements, and data sovereignty issues. Federated Learning \cite{li2020federated, mcmahan2017communication, zhang2021survey} has emerged as a promising solution, enabling collaborative model training while keeping sensitive data localized.
Fine-tuning foundation models like LLMs and Vision Transformers \cite{dosovitskiy2021an} presents significant computational challenges, requiring substantial GPU resources and memory. Parameter-efficient fine-tuning (PEFT) methods \cite{pmlr-v97-houlsby19a}, particularly Low-Rank Adaptation (LoRA) \cite{hu2021lora}, have gained widespread adoption by reducing the number of trainable parameters while maintaining performance. However, recent theoretical analyses and empirical studies, notably FedFTG \cite{mahla2025exploringgradientsubspacesaddressing}, have revealed fundamental limitations in LoRA's effectiveness within federated settings, primarily due to constrained gradient subspaces that limit model expressiveness during distributed training.
Our work makes several key contributions to address these challenges:
\begin{itemize}
    \item We propose a novel parameter-efficient architecture for federated fine-tuning that introduces a compressed multi-layer perceptron (MLP) between the up-projection and down-projection layers in the transformer's feed-forward network. This design enables more expressive parameter updates while maintaining computational efficiency.
    \item Through extensive empirical analysis, we demonstrate that our method consistently outperforms recent approaches including FFA-LoRA \citep{sun2024improving} and FedSA-LoRA \cite{guo2024selectiveaggregationlowrankadaptation} across both vision and language tasks. While FFA-LoRA incorporates differential privacy mechanisms, our primary focus is on evaluating their fundamental aggregation strategies and model adaptation techniques. We adopt their non-private variants (ie. we perform the experiments in the same way as was done in the experiments of the FFA-LoRA paper without considering differential privacy) as baselines to ensure a fair comparison of architectural innovations, isolating the impact of privacy mechanisms from core model performance.
    \item We provide theoretical insights into why our approach achieves superior convergence compared to LoRA-based methods in federated settings, supported by empirical evidence across diverse tasks and data distributions.
\end{itemize}

% Our experiments demonstrate that the proposed architecture achieves state-of-the-art performance while maintaining practical efficiency for real-world deployments. The method's simplicity and effectiveness make it particularly suitable for scenarios where computational resources are limited and data privacy is paramount.

\section{Related Works}
\subsection{Federated Learning}
Federated Learning (FL) is a widely used distributed learning approach for privacy-sensitive tasks, but it faces significant challenges when working with non-IID datasets, often resulting in accuracy gaps compared to centralized training. Recent research has shown that fine-tuning Pre-trained Language Models (PLMs) in FL can help mitigate these challenges \citep{nguyen2022begin}, with vanilla FedAvg achieving performance on par with centralized methods. Foundational models like Large Language Models (LLMs) and Vision Transformers (ViTs), which need to be trained in privacy-preserving environments, are particularly well-suited for FL. However, fine-tuning these models remains difficult due to their high computational demands, driven by their large number of parameters. To address these resource limitations, Parameter-Efficient Fine-Tuning (PEFT) methods such as adapter-tuning, prompt-tuning, and LoRA have been introduced. Among these, LoRA-based techniques have gained widespread attention for their efficiency, effectiveness, and flexibility, which is the central focus of our work.

\subsection{LoRA in Federated Learning}
Low-Rank Adaptation (LoRA) \citep{hu2021lora} utilizes low-rank matrices to approximate gradient updates, enabling the adaptation of models without altering pre-trained weights. This method has gained popularity due to its efficiency, effectiveness, and adaptability. By significantly reducing the size of model updates, LoRA helps address communication bottlenecks in Federated Learning (FL). For instance, \citet{yi2023fedlora} introduced FedLoRA, which integrates LoRA into FL to improve fine-tuning efficiency. Similarly, \citet{yang2024dual} proposed FedDPA, a dual-adapter approach, and \citet{qi2024fdlora} presented FDLoRA, both leveraging personalized and global LoRA modules to capture local and shared knowledge. Federated Freeze-A LoRA (FFA-LoRA) \citep{sun2024improving} freezes the randomly initialized A matrices and only fine-tuning the zero-initialized B matrices, further halving communication costs and showing promising performance improvements compared to the other LoRA baselines. FedSA-LoRA \cite{guo2024selectiveaggregationlowrankadaptation} takes a different approach by fine-tuning both matrices but selectively aggregating only the randomly initialized matrix A. FedFMSL \citep{10666083} introduces a two-stage approach combining global and local experts with a novel Sparsely Activated LoRA (SAL) algorithm. LoRA-FAIR \citep{bian2024lorafairfederatedlorafinetuning} addresses two fundamental challenges in federated LoRA: server-side aggregation bias and client-side initialization drift. It introduces a correction term at the server while maintaining the original LoRA modules, demonstrating consistent performance improvements across Vision Transformer and MLP-Mixer architectures. Focusing on communication efficiency, \citet{kuo2024federatedlorasparsecommunication} propose FLASC, which applies sparsity during communication while allowing full local LoRA fine-tuning. Their approach achieves performance comparable to dense LoRA with up to 10× reduction in communication costs, while simultaneously showing benefits for heterogeneity and privacy concerns. 

% Heterogeneous LoRA explores using different ranks across clients, with aggregation performed via zero-padding and redistribution through truncation. However, this zero-padding approach often causes instability during training. To counter this, \citet{byun2024towards} proposed a replication-based aggregation strategy for rank-heterogeneous LoRA. Additionally, \citet{qiu2024federated} introduced Rank-Based LoRA Aggregation (RBLA), which uses weighted aggregation for different LoRA structures, while \citet{wang2024flora} developed a stacking-based aggregation method.

\section{Methodology}
\begin{figure}[h]
  \centering
  \includegraphics[width=0.34\textwidth]{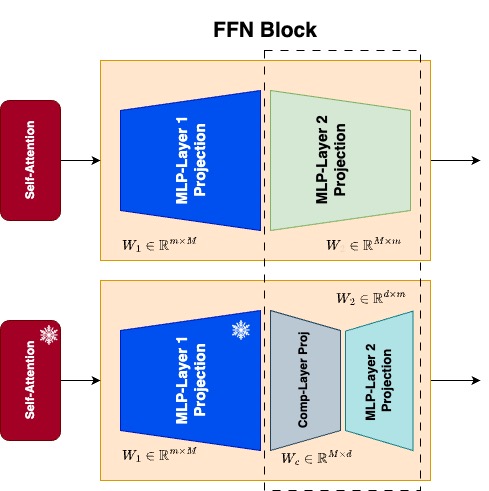}
  \caption{The block at the top represents the original architecture and the one at the bottom represents the architecture with compression layers added. MLP Layer 1 and Self Attention modules are frozen during training. The new compression layer is added between MLP Layer 1 and MLP Layer 2 such that $d\ll M,m$}
  \label{fig:fedftgplus}
\end{figure}
Recent studies have provided both theoretical analysis and empirical evidence highlighting the limitations of using LoRA for federated fine-tuning of foundational models, primarily due to its constrained subspace learning \cite{mahla2025exploringgradientsubspacesaddressing}. Building on these insights, we propose a novel approach to federated fine-tuning that does not rely on LoRA. Instead, our method employs direct weight averaging using FedAvg, without the need for low-rank parallel adapters, while maintaining parameter efficiency comparable to LoRA. As illustrated in the figure, we introduce a much smaller MLP layer immediately after the MLP layer that follows the self-attention module. This results in a modification of the projection in MLP layer 2, as depicted in the figure. This additional intermediate layer is inspired by observations from \cite{tian2024joma}, which suggest that explicit fine-tuning of the self-attention module is unnecessary, as its information is already captured by the MLP layer directly following it. Consequently, we apply the compression layer right after the first MLP layer in the transformer block. By doing so, we manipulate the information passing through the attention module, projecting it into a smaller subspace and learning representations within this reduced dimension similar to what has been previously discussed in \cite{mahla2025exploringgradientsubspacesaddressing}. During training, only the parameters of the compression layer ($\boldsymbol{W_c}$) and the projection following it ($\boldsymbol{W_2}$) are fine-tuned. This way, the number of parameters fine-tuned are $d\times(M+m)$. Clearly, having a compression layer having output dimension $d$ is analogous to having a LoRA adapter with rank $d$ but parallel to the pre-trained weight matrix showcasing the parameter efficiency that our approach brings to the table with better performance guarantees. \newline
We randomly initialize $\boldsymbol{W_c}$ with a unit Gaussian distribution. For any output value $\boldsymbol{x}$ of the MLP-Layer 1 Projection, the following relation can be written assuming linear activation between the projection layers:
\begin{equation*}
    \boldsymbol{W_2}\boldsymbol{W_c} = \boldsymbol{W}
\end{equation*}
\begin{equation}
    \Rightarrow \boldsymbol{W_2} = \boldsymbol{W}\boldsymbol{W_c^{+}}
\end{equation}
Where $\boldsymbol{W_c^{+}}$ is the pseudoinverse (Moore-Penrose inverse) of $\boldsymbol{W_c}$.
Equation (1) represents how the weights of the modified MLP-2 Layer are initialized. \newline
\textbf{Lemma 1:}\textit{For a bounded gradient (L2 norm of the gradients upper bounded by $D$) L2 norm of the weight matrix in the sequential compression layer based FedAvg aggregation framework is upper bounded linearly by the number of global aggregation steps $S$ and the number of local training steps between two consecutive aggregation steps $t_{agg}$:
    \begin{equation}
        \left\| \boldsymbol{W_{agg}} \right\| \le B + \eta St_{agg}D = \mathcal{O}(St_{agg})
    \end{equation}
    where $\eta$ is the learning rate and B is a constant. }\newline

% \textbf{Proof:} Weight update for a client $i$ at some time $t_{agg}$ with FedAvg aggregation and $S$ communication rounds being occurred, can be written as:

% \begin{equation}
%     \boldsymbol{W_{agg}}=\boldsymbol{W_0} - \frac{\eta}{N}\sum_{i=1}^N\sum_{j=0}^{S}\sum_{t=0}^{t_{agg}}\boldsymbol{G_{t,j}^{(i)}}
% \end{equation}
% Here, $\boldsymbol{G}$ is the gradient. Taking L2 norm on both the sides: 
% \begin{equation*}
%     \left\| \boldsymbol{W_{agg}} \right\| \le \left\| \boldsymbol{W_0} \right\| + \frac{\eta}{N}\sum_{i=1}^N\sum_{j=0}^S\sum_{t=0}^{t_{agg}}D 
% \end{equation*}
% \begin{equation}
%     \implies \left\| \boldsymbol{W_{agg}} \right\|\le B + \eta S t_{agg}D
% \end{equation}
Following this Lemma, we propose the following theorem. \newline
\newline
\textbf{Theorem 1:}
    \textit{For a convex loss $\mathcal{L}$, let $\boldsymbol{\Delta W^*} \in \mathbb{R}^{d \times k}$ ($r \ll \text{min}(d,k)$) be the optimal parameter matrix, $\alpha$ be the learning rate, and let the L2 norm of the gradient to be bounded (i.e. $\left\| \boldsymbol{\nabla_W \mathcal{L}^{(i)}(\Delta W)} \right\|_2 \le D$). The excess risk ($\left| \mathcal{L}(\boldsymbol{\Delta W_{agg}}) - \mathcal{L}(\boldsymbol{\Delta W^*}) \right|$) bounds for our method, involving $N$ clients and $S$ global aggregation steps having occurred every $t_{agg}$ local training iterations, can be expressed as follows: 
    \begin{equation}
   \left| \mathcal{L}(\boldsymbol{\Delta W_{agg}}) - \mathcal{L}(\boldsymbol{\Delta W^*}) \right| \le \alpha D^2St_{agg}+c = \mathcal{O}(St_{agg})
\end{equation}}\newline
% \textbf{Proof:} \newline
% \begin{equation}
%     \mathcal{L}(\boldsymbol{\Delta W}) - \mathcal{L}(\boldsymbol{\Delta W^*}) \le \boldsymbol{\nabla_W \mathcal{L}}(\boldsymbol{\Delta W})^\top(\boldsymbol{\Delta W}-\boldsymbol{\Delta W}^*)
% \end{equation}
% For the excess risk just after the aggregation, we can replace the weight parameters with the average of it.
% \begin{equation}
%     \mathcal{L}(\boldsymbol{\Delta W_{agg}}) - \mathcal{L}(\boldsymbol{\Delta W^*}) \le \boldsymbol{\nabla_W \mathcal{L}}(\boldsymbol{\Delta W_{agg}})^\top(\boldsymbol{\Delta W_{agg}}-\boldsymbol{\Delta W}^*)
% \end{equation}
% Taking L2 norm on both the sides and using the bounds of $\boldsymbol{W_{agg}}$ from Lemma 1:
% \begin{equation*}
%     \left| \mathcal{L}(\boldsymbol{\Delta W_{agg}}) - \mathcal{L}(\boldsymbol{\Delta W^*}) \right| \le D (\alpha t_{agg} SD + k)= \mathcal{O}(St_{agg})
% \end{equation*}

Theorem 1 demonstrates our method achieves linear excess risk bounds, independent of the number of clients $N$, while LoRA methods have quadratic bounds that scale with $N$ \cite{mahla2025exploringgradientsubspacesaddressing}. This advantage stems from our sequential compression design, which creates a direct path for gradient flow through a controlled subspace, preventing error accumulation during federated averaging. In contrast, LoRA's parallel adapters can lead to divergent updates across clients, causing quadratic error accumulation.

\section{Experiments}
In this section, we evaluate the performance of our method across text and image modalities, focusing on its performance with LLMs and ViTs. For text, experiments involve 3 and 4 clients, while for vision, 3, 4, and 5 clients are tested. Each client holds a unique non-IID shard of a dataset created by partitioning the original dataset using Dirichlet Allocation with concentration parameter $\alpha$ equal to 0.1.

\subsection{Datasets}
We conduct experiments on both text and image datasets. For text, we use the MedQuAD dataset \cite{BenAbacha-BMC-2019}, which includes 47,457 medical question-answer pairs from 12 NIH websites, covering 39 question types. Due to MedlinePlus copyright restrictions, answers from 3 subsets were removed. We also use the Databricks Dolly-15k dataset \cite{DatabricksBlog2023DollyV2}, which contains 15,000 high-quality human-generated prompt-response pairs for instruction tuning LLMs, with categories like brainstorming, classification, summarization, and question answering. For the vision modality, we use the Brain Tumor classification dataset \cite{https://doi.org/10.6084/m9.figshare.1512427.v5}, comprising 3,064 T1-weighted contrast-enhanced MRI images from 233 patients, categorized into meningioma, glioma, and pituitary tumor types.

\subsection{Non-IID Data Preparation}
\begin{figure}[h]
  \centering
  \includegraphics[width=8cm]{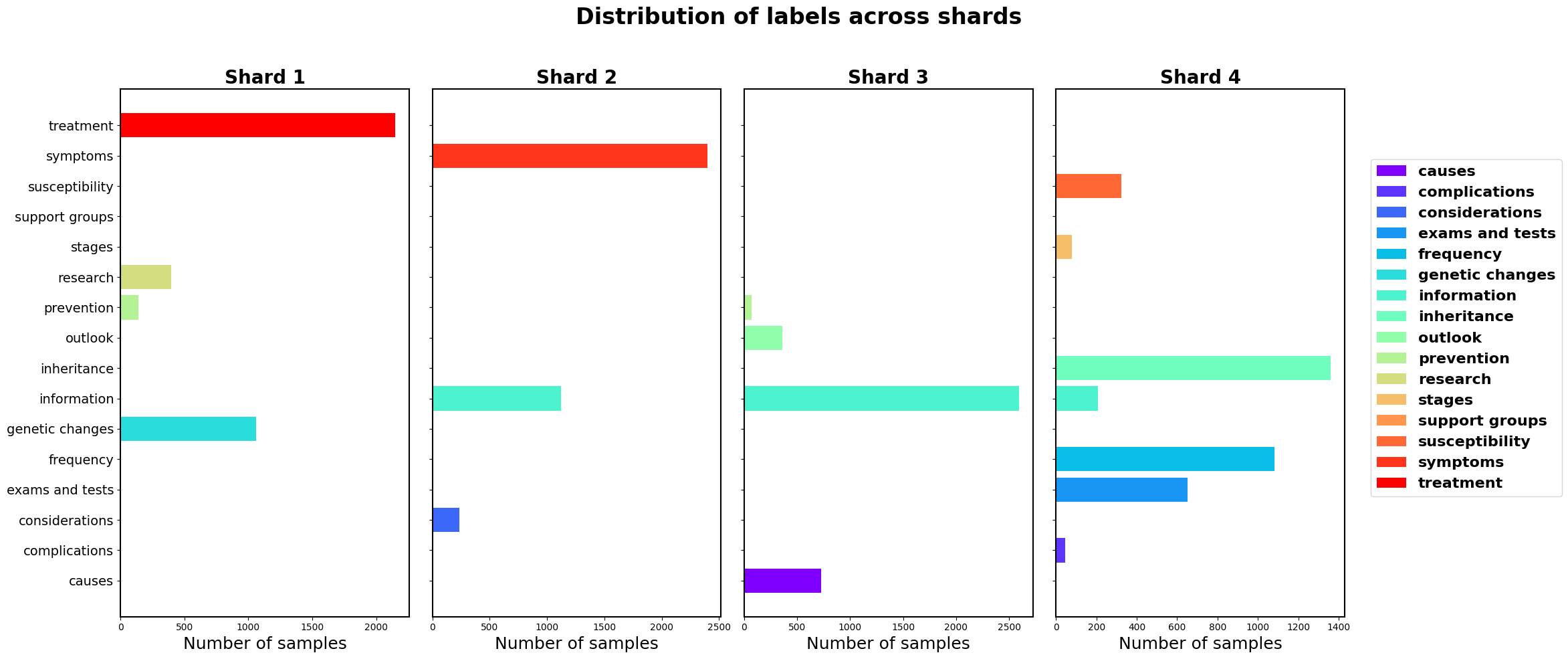}
  \caption{From \cite{mahla2025exploringgradientsubspacesaddressing}: Label distribution across shards for the MedQuAD dataset produced using Dirichlet Allocation with $\alpha=0.1$.}
  \label{fig:shard-medquad}
\end{figure}

\begin{figure}[h]
  \centering
  \includegraphics[width=8cm]{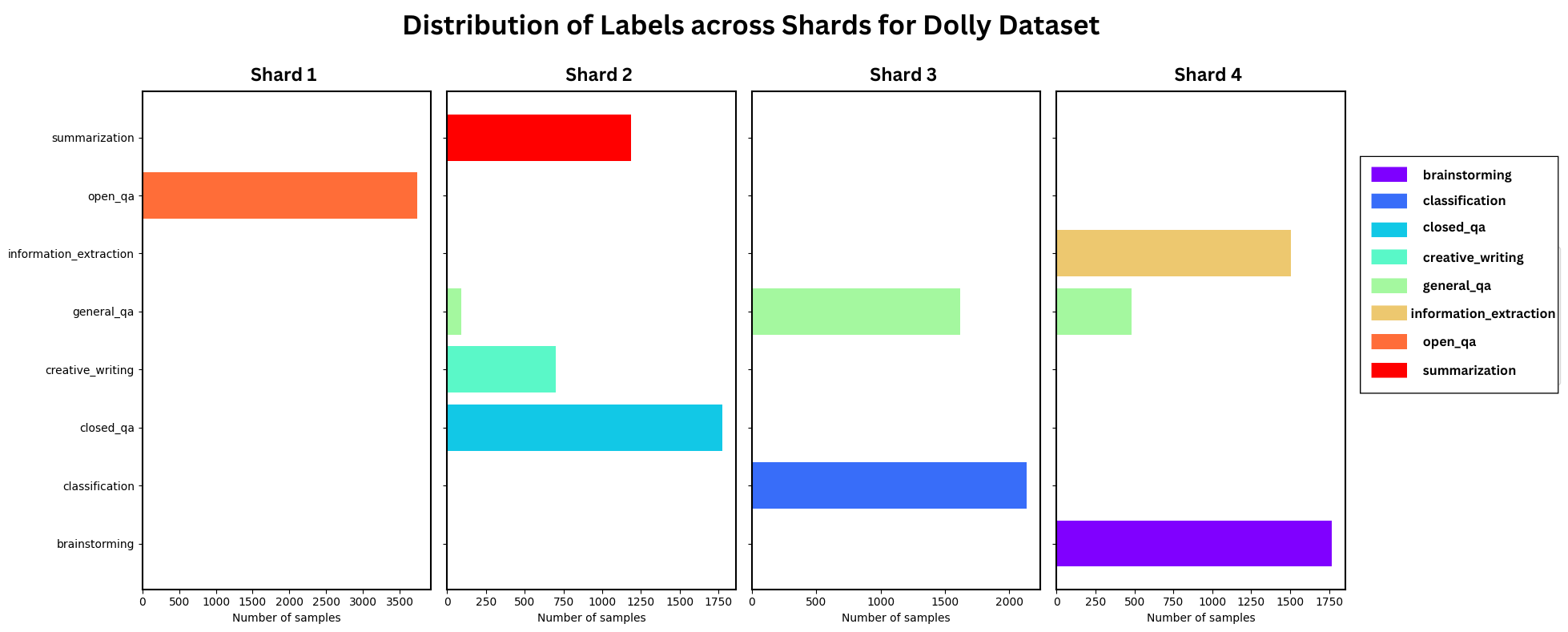}
  \caption{From \cite{mahla2025exploringgradientsubspacesaddressing}: Label distribution across shards for the Dolly dataset produced using Dirichlet Allocation with $\alpha=0.1$.}
  \label{fig:shard-dolly}
\end{figure}

To simulate non-IID conditions, we used Dirichlet Allocation to partition each dataset into non-IID splits, following \cite{2305.05644, sun2024improving}. For text datasets (MedQuAD and Dolly-15k), we created 4 splits, and for the Brain Tumor dataset, 5 splits. The concentration parameter $\alpha$ was set to 0.1, resulting in highly skewed distributions across clients. Splits were based on labels: \textit{question\_type} for MedQuAD and \textit{category} for Dolly-15k. Class label distributions for MedQuAD and Dolly-15k are shown in Figure \ref{fig:shard-medquad} and \ref{fig:shard-dolly} respectively. A similar class label distribution for the Brain Tumor Classification dataset is shown in the appendix.

\subsection{Models and Hyperparameters}
For experiments on text datasets (MedQuAD and Dolly-15k), we use Gemma-2B \cite{gemmateam2024gemmaopenmodelsbased} and Tinyllama-1.1B \cite{zhang2024tinyllamaopensourcesmalllanguage}. Following our methodology, we fine-tune two components: the compression-layer projection weights and the modified MLP-layer 2 weights. The compression layer dimension was set to 8, which maintains parameter efficiency comparable to LoRA while ensuring sufficient expressivity for learning task-specific adaptations. This choice results in $d \times (M+m)$ trainable parameters, where $d=8$ is the compression dimension and $M,m$ are the original MLP layer dimensions. For fair comparison, we implement FedSA-LoRA and FFA-LoRA with matching parameter counts, using a LoRA rank of 8 and a scaling factor of 16 for attention module fine-tuning (query, key, value), following their original implementations.
For vision experiments using SigLIP \cite{2303.15343}, we maintain architectural consistency by applying the same compression dimension ($d=8$) while additionally fine-tuning the classifier layer to accommodate vision-specific requirements. The baseline LoRA methods (FedSA-LoRA and FFA-LoRA) were configured with rank 8 and a scaling factor of 32, adjusted for vision tasks while maintaining comparable parameter counts to our approach.

% \begin{figure}[h]
%   \centering
%   \includegraphics[width=0.85\linewidth]{figures/vision_shards.jpeg}
%   \caption{From \cite{mahla2025exploringgradientsubspacesaddressing}: Distribution of labels across shards for Brain Tumour Dataset produced using Dirichlet Allocation with $\alpha=0.1$}
%   \label{fig:shard-vision}
% \end{figure}

\subsection{Experiment Results}
Tables 1 and 2 show results on text and image modality respectively. We train the models for 10 epochs with early stopping owing to overfitting. We report the ROUGE\_L (ROUGE longest common subsequence) score and BLUE-4 (4-gram) score for text modalities while F1 score for the experiments on image modality.

\begin{table}[]
\centering
\begin{tabular}{l|l|l|l|l|l}
\hline
$N$                 & Dataset                    & Model                      & Method   & BLEU-4          & ROUGE-L         \\ \hline
\multirow{12}{*}{3} & \multirow{6}{*}{MedQuAD}   & \multirow{3}{*}{TinyLlama} & Ours   & \textbf{0.3204} & \textbf{0.4576} \\
                    &                            &                            & FedSA-LoRA & 0.1150          & 0.3071          \\
                    &                            &                            & FFA-LoRA & 0.1004          & 0.2947          \\ \cline{3-6} 
                    &                            & \multirow{3}{*}{Gemma-2B}  & Ours   & \textbf{0.3734} & \textbf{0.5370} \\
                    &                            &                            & FedSA-LoRA & 0.1209          & 0.3065          \\
                    &                            &                            & FFA-LoRA & 0.1077          & 0.2875          \\ \cline{2-6} 
                    & \multirow{6}{*}{Dolly-15K} & \multirow{3}{*}{TinyLlama} & Ours   & \textbf{0.4133} & \textbf{0.5107} \\
                    &                            &                            & FedSA-LoRA & 0.0248          & 0.0848          \\
                    &                            &                            & FFA-LoRA & 0.0566          & 0.1708          \\ \cline{3-6} 
                    &                            & \multirow{3}{*}{Gemma-2B}  & Ours   & \textbf{0.3915} & \textbf{0.5227} \\
                    &                            &                            & FedSA-LoRA & 0.0786          & 0.0872          \\
                    &                            &                            & FFA-LoRA & 0.1077          & 0.2875          \\ \hline
\multirow{12}{*}{4} & \multirow{6}{*}{MedQuAD}   & \multirow{3}{*}{TinyLlama} & Ours   & \textbf{0.3283} & \textbf{0.4903} \\
                    &                            &                            & FedSA-LoRA & 0.1164           & 0.2953          \\
                    &                            &                            & FFA-LoRA & 0.1133          & 0.3047          \\ \cline{3-6} 
                    &                            & \multirow{3}{*}{Gemma-2B}  & Ours   & \textbf{0.3634} & \textbf{0.5435} \\
                    &                            &                            & FedSA-LoRA & 0.1130          & 0.2842          \\
                    &                            &                            & FFA-LoRA & 0.1092          & 0.2863          \\ \cline{2-6} 
                    & \multirow{6}{*}{Dolly-15K} & \multirow{3}{*}{TinyLlama} & Ours   & \textbf{0.3516} & \textbf{0.4704} \\
                    &                            &                            & FedSA-LoRA & 0.0363          & 0.0952          \\
                    &                            &                            & FFA-LoRA & 0.0619          & 0.1665          \\ \cline{3-6} 
                    &                            & \multirow{3}{*}{Gemma-2B}  & Ours   & \textbf{0.3511} & \textbf{0.5233} \\
                    &                            &                            & FedSA-LoRA & 0.0346          & 0.0984          \\
                    &                            &                            & FFA-LoRA & 0.0661          & 0.1648          \\ \hline
\end{tabular}
\caption{Comparison of BLEU-4 and ROUGE L F1 scores across different methods, models, and datasets for varying client numbers with non-IID splits (see fig. \ref{fig:shard-dolly} and \ref{fig:shard-medquad})}
\label{tab1:results-main}
\end{table}
As demonstrated in table \ref{tab1:results-main}, our sequential compression layer approach significantly outperforms recent state-of-the-art LoRA methods (FedSA-LoRA and FFA-LoRA) across text-based tasks. Similar performance gains are observed in vision modality experiments (table \ref{tab2:results-vision}). These results empirically validate our theoretical analysis - while LoRA-based methods suffer from quadratic excess risk bounds that scale with both the number of clients and aggregation steps, our method's linear bounds ($\mathcal{O}(St_{agg})$) enable more stable learning. The strategic placement of our compression layer between MLP layers, rather than parallel adapters as in LoRA, allows for more expressive parameter updates within a controlled subspace, leading to better capture of task-specific semantics across both modalities while maintaining comparable parameter efficiency.

\begin{table}[]
\centering
\begin{tabular}{lll}
\hline
No of clients      & Method   & F1 Score        \\ \hline
\multirow{3}{*}{3} & Ours   & \textbf{0.481} \\
                   & FedSA-LoRA & 0.475          \\
                   & FFA-LoRA & 0.0165           \\ \hline
\multirow{3}{*}{4} & Ours   & \textbf{0.577}  \\
                   & FedSA-LoRA & 0.315          \\
                   & FFA-LoRA & 0.3175          \\ \hline
\multirow{3}{*}{5} & Ours   & \textbf{0.516} \\
                   & FedSA-LoRA & 0.320          \\
                   & FFA-LoRA & 0.1104         
\end{tabular}
\caption{Comparison of F1 score on Brain Tumour Classification Dataset for fine-tuning SigLIP using different FL fine-tuning methods (see figure in appendix}
\label{tab2:results-vision}
\end{table}

\section{Conclusion}
This paper presents a novel approach to federated fine-tuning of foundational models through strategically placed sequential compression layers, offering both theoretical and practical advantages over existing LoRA-based methods. Our key innovation lies in projecting representations into a carefully designed compressed subspace immediately after the self-attention module, enabling more expressive parameter updates while maintaining efficiency. Unlike LoRA-based approaches that suffer from constrained subspace learning and quadratic excess risk bounds, our method achieves linear bounds on both weight updates and excess risk, independent of the number of clients. This theoretical advantage translates to practical benefits, as demonstrated through extensive experiments across both language (Gemma-2B, TinyLlama-1.1B) and vision (SigLIP) models under challenging non-IID data distributions.
The empirical results consistently show superior performance over state-of-the-art methods like FFA-LoRA and FedSA-LoRA while maintaining comparable parameter efficiency. Our approach's ability to achieve better results makes it particularly suitable for real-world federated learning deployments where resources are constrained and data privacy is crucial. Furthermore, the method's simplicity and theoretical guarantees provide a solid foundation for future research in efficient, privacy-preserving distributed learning of foundation models.

\bibliography{iclr2025_conference}
\bibliographystyle{iclr2025_conference}

\appendix
\section{Appendix}
\subsection{Proofs}
\textbf{Lemma 1:}\textit{For a bounded gradient (L2 norm of the gradients upper bounded by $D$) L2 norm of the weight matrix in the sequential compression layer based FedAvg aggregation framework is upper bounded linearly by the number of global aggregation steps $S$ and the number of local training steps between two consecutive aggregation steps $t_{agg}$:
    \begin{equation}
        \left\| \boldsymbol{W_{agg}} \right\| \le B + \eta St_{agg}D = \mathcal{O}(St_{agg})
    \end{equation}
    where $\eta$ is the learning rate and B is a constant. }\newline

\textbf{Proof:} Weight update for a client $i$ at some time $t_{agg}$ with FedAvg aggregation and $S$ communication rounds being occurred, can be written as:

\begin{equation}
    \boldsymbol{W_{agg}}=\boldsymbol{W_0} - \frac{\eta}{N}\sum_{i=1}^N\sum_{j=0}^{S}\sum_{t=0}^{t_{agg}}\boldsymbol{G_{t,j}^{(i)}}
\end{equation}
Here, $\boldsymbol{G}$ is the gradient. Taking L2 norm on both the sides: 
\begin{equation*}
    \left\| \boldsymbol{W_{agg}} \right\| \le \left\| \boldsymbol{W_0} \right\| + \frac{\eta}{N}\sum_{i=1}^N\sum_{j=0}^S\sum_{t=0}^{t_{agg}}D 
\end{equation*}
\begin{equation}
    \implies \left\| \boldsymbol{W_{agg}} \right\|\le B + \eta S t_{agg}D
\end{equation}
\newline
\newline
\textbf{Theorem 1:}
    \textit{For a convex loss $\mathcal{L}$, let $\boldsymbol{\Delta W^*} \in \mathbb{R}^{d \times k}$ ($r \ll \text{min}(d,k)$) be the optimal parameter matrix, $\alpha$ be the learning rate, and let the L2 norm of the gradient to be bounded (i.e. $\left\| \boldsymbol{\nabla_W \mathcal{L}^{(i)}(\Delta W)} \right\|_2 \le D$). The excess risk ($\left| \mathcal{L}(\boldsymbol{\Delta W_{agg}}) - \mathcal{L}(\boldsymbol{\Delta W^*}) \right|$) bounds for our method, involving $N$ clients and $S$ global aggregation steps having occurred every $t_{agg}$ local training iterations, can be expressed as follows: 
    \begin{equation}
   \left| \mathcal{L}(\boldsymbol{\Delta W_{agg}}) - \mathcal{L}(\boldsymbol{\Delta W^*}) \right| \le \alpha D^2St_{agg}+c = \mathcal{O}(St_{agg})
\end{equation}}\newline
\textbf{Proof:} \newline
\begin{equation}
    \mathcal{L}(\boldsymbol{\Delta W}) - \mathcal{L}(\boldsymbol{\Delta W^*}) \le \boldsymbol{\nabla_W \mathcal{L}}(\boldsymbol{\Delta W})^\top(\boldsymbol{\Delta W}-\boldsymbol{\Delta W}^*)
\end{equation}
For the excess risk just after the aggregation, we can replace the weight parameters with the average of it.
\begin{equation}
    \mathcal{L}(\boldsymbol{\Delta W_{agg}}) - \mathcal{L}(\boldsymbol{\Delta W^*}) \le \boldsymbol{\nabla_W \mathcal{L}}(\boldsymbol{\Delta W_{agg}})^\top(\boldsymbol{\Delta W_{agg}}-\boldsymbol{\Delta W}^*)
\end{equation}

Taking L2 norm on both the sides and using the bounds of $\boldsymbol{W_{agg}}$ from Lemma 1:
\begin{equation*}
    \left| \mathcal{L}(\boldsymbol{\Delta W_{agg}}) - \mathcal{L}(\boldsymbol{\Delta W^*}) \right| \le D (\alpha t_{agg} SD + k)= \mathcal{O}(St_{agg})
\end{equation*}

\subsection{Dataset Shard Class Distribution}

Below figure shows the class label distributions for the Brain Tumor Classification Dataset. The dataset is sharded using Dirichlet Allocation with a concentration parameter $\alpha=0.1$. 

\begin{figure}[h]
  \centering
  \includegraphics[width=0.85\linewidth]{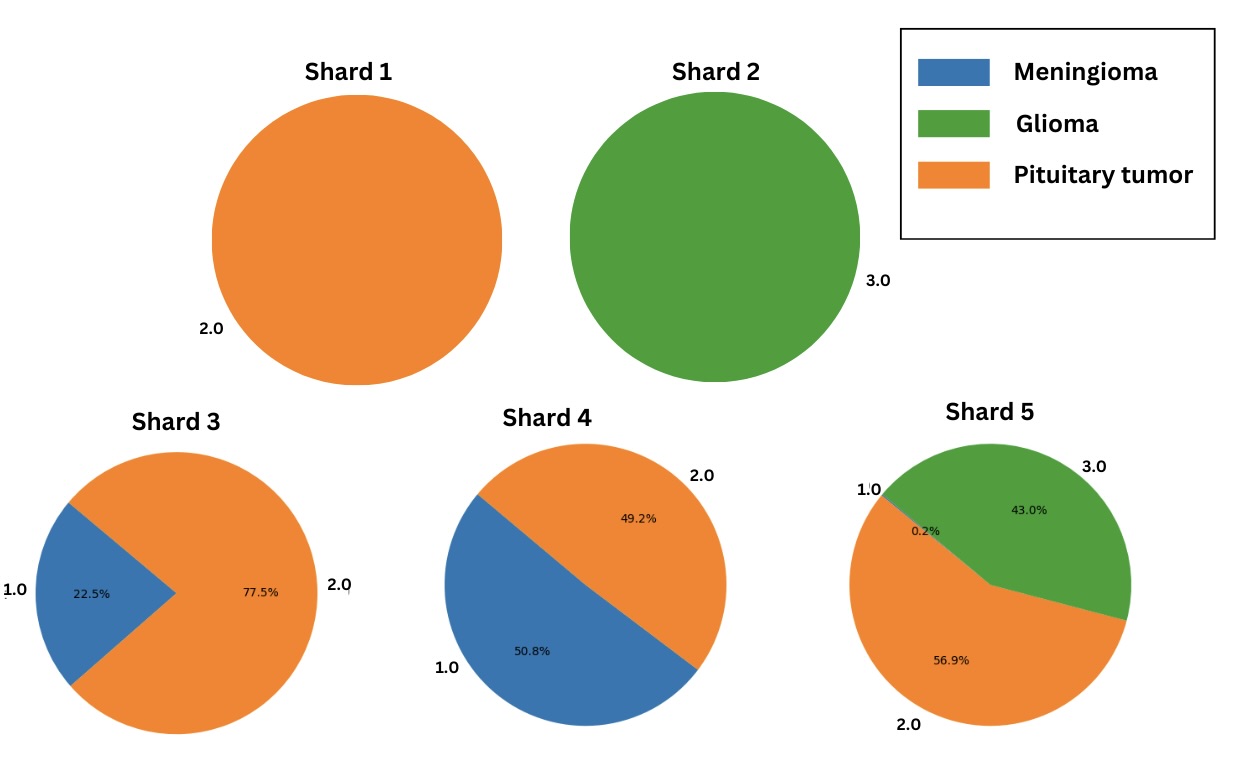}
  \caption{From \cite{mahla2025exploringgradientsubspacesaddressing}: Distribution of labels across shards for Brain Tumour Dataset produced using Dirichlet Allocation with $\alpha=0.1$}
  \label{fig:shard-vision}
\end{figure}

\end{document}

%% file: math_commands.tex
\usepackage{amsmath,amsfonts,bm}

\def\eqref#1{equation~\ref{#1}}
\def\1{\bm{1}}

\DeclareMathAlphabet{\mathsfit}{\encodingdefault}{\sfdefault}{m}{sl}
\SetMathAlphabet{\mathsfit}{bold}{\encodingdefault}{\sfdefault}{bx}{n}